\documentclass[conference]{IEEEtran}
\IEEEoverridecommandlockouts
\usepackage{cite}
\usepackage{amsmath,amssymb,amsfonts}
\usepackage{algorithmic}
\usepackage{graphicx}
\usepackage{textcomp}
\usepackage{xcolor}
\begin{document}

\title{Phase Space Reconstruction Network for \\ Lane Intrusion Action Recognition 
\thanks{\textsuperscript{$\ast$} Corresponding author: Zhidong Deng.}
}
\author{\IEEEauthorblockN{Ruiwen Zhang, \textsuperscript{$\ast$}Zhidong Deng, Hongsen Lin, Hongchao Lu}
\IEEEauthorblockA{
\textit{Institute for Artificial Intelligence at Tsinghua University (THUAI)} \\
\textit{State Key Laboratory of Intelligent Technology and Systems}\\
\textit{Beijing National Research Center for Information Science and Technology (BNRist)} \\
\textit{Center for Intelligent Connected Vehicles and Transportation} \\
\textit{Department of Computer Sicence and Technology, Tsinghua University,} Beijing 100084, China \\
\{zhangrw18, lin-hs20, luhc15\}@mails.tsinghua.edu.cn, \textsuperscript{$\ast$}michael@mail.tsinghua.edu.cn}
}
\maketitle

\begin{abstract}
In a complex road traffic scene, illegal lane intrusion of pedestrians or cyclists constitutes one of the main safety challenges in autonomous driving application. In this paper, we propose a novel object-level phase space reconstruction network (PSRNet) for motion time series classification, aiming to recognize lane intrusion actions that occur 150m ahead through a monocular camera fixed on moving vehicle. 
In the PSRNet, the movement of pedestrians and cyclists, specifically viewed as an observable object-level dynamic process, can be reconstructed as trajectories of state vectors in a latent phase space and further characterized by a learnable Lyapunov exponent-like classifier that indicates discrimination in terms of average exponential divergence of state trajectories. 
Additionally, in order to first transform video inputs into one-dimensional motion time series of each object, a lane width normalization based on visual object tracking-by-detection is presented. 
Extensive experiments are conducted on the THU-IntrudBehavior dataset collected from real urban roads. The results show that our PSRNet could reach the best accuracy of 98.0\%, which remarkably exceeds existing action recognition approaches by more than 30\%.
\end{abstract}


\section{Introduction}
\begin{figure}[t]
    \centering
    \includegraphics[width=0.9\columnwidth]{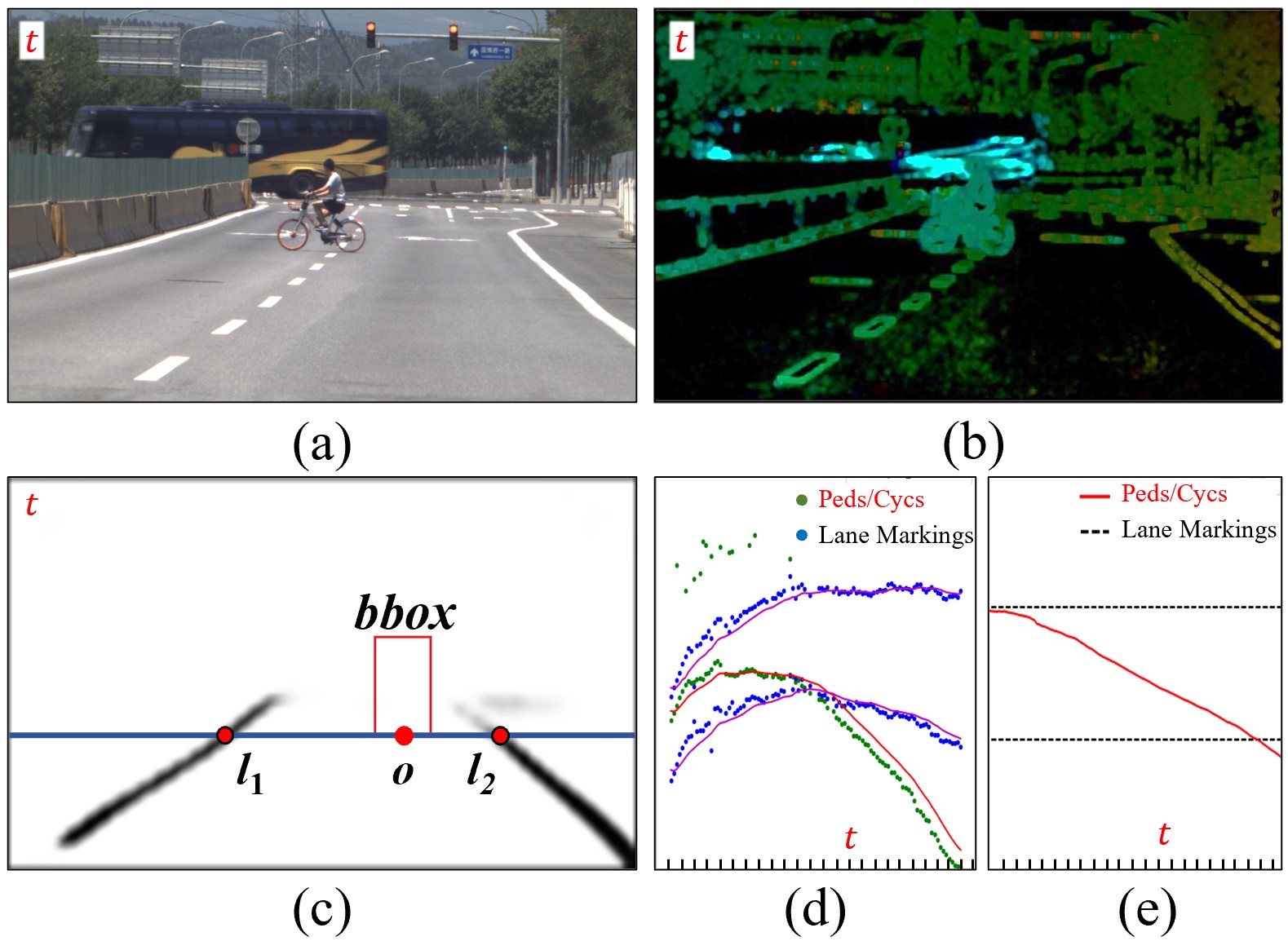}
    \caption{(a) Raw video frames. (b) Optical flow visualization of raw frames. (c) Visual object detection results in raw frames. Lane intrusion actions are only dependent on positions of moving objects $o$ (i.e., pedestrians and cyclists) relative to central trajectory of current lanes constituted by $l_1$ and $l_2$. (d) Resulting motion time series of each object. (e) Normalized relative motion time series.}
    \label{introduce}
    \vspace{-0.5cm}
\end{figure}

In recent years, autonomous driving has become a hot and promising area in industry and academic circles. 
Although most of moving objects are motor vehicles in highway scenarios, 
there occasionally exist some unusual emergencies, including illegal lane intrusion of pedestrians or cyclists, 
which actually constitutes one of the main safety challenges in autonomous driving application. 
Focusing on this problem, we aim to recognize lane intrusion actions that occur at least 150m ahead solely based on a monocular camera fixed on vehicles moving at high speed. 

Currently, there are many trajectory prediction methods and datasets~\cite{gupta2018social,zhu2019starnet,sun2020recursive,pellegrini2009you,lerner2007crowds} based on fixed camera. However, those methods usually do not work well while camera is moving. It is hard to determine relative location of objects because background changes for moving camera. Video action recognition models~\cite{karpathy2014large,simonyan2014two,lin2019tsm,feichtenhofer2019slowfast,tran2018closer,zhou2018temporal,yang2020temporal} can simultaneously have extraction of spatial and temporal features in video frames and are also exploited to recognize actions. But unfortunately, the above-mentioned approaches are often well suited to low-speed and close-range situations, which implies that camera is fixed or moves slowly, and particularly moving object often occupies a large proportion of a pixel plane in a video frame. 

This paper mainly investigates high-speed and long-range scenarios in a structured road traffic environment. In such a situation, on-board camera moves forward at high speed with ego-vehicle, and moving objects ahead on road usually occur about 150-250m away in front of camera. Undoubtedly, it will bring new big challenges, such as inconsistent visual changes of moving objects in different video frames and varying view-angles caused by vehicle steering. Interestingly, it is readily observed from Fig.~\ref{introduce}(b) that optical flow may represent the motion of vehicles instead of cyclists. All the challenges mean that pixel changes in consecutive video frames are likely to be unable to truly reflect actual motion of all objects. Owing to the fact that moving objects frequently occupy a small proportion of a pixel plane in a single video frame due to being far from camera, it is very difficult to detect such tiny moving objects and recognize their actions accurately. From this perspective, most of the methods proposed in the past are tough to solve intrusion action recognition problems.

For high-speed and long-range scenes, this paper first presents an object-level lane width normalization strategy to transform lane intrusion actions of objects including pedestrians or cyclists in video frames into relative motion time series of objects. Through normalizing relative positions of objects with pixel-level lane widths at current video frame, we can thus give rise to one-dimensional motion time series of objects, which could be employed to address the challenge caused by moving camera, i.e., pixel changes in video frames may be unable to reflect actual actions of objects. Second, we propose a PSRNet to classify the above motion time series and then eventually complete lane intrusion action recognitions. Inspired by prediction methods of phase-space reconstruction, we make use of several convolutional layers in the PSRNet to reconstruct latent phase spaces of dynamic system embedded in observable motion time series, which actually expresses the motion of object trajectories in a latent state space. Following this, we set up a learnable Lyapunov exponent-like classifier in our PSRNet, in order to learn invariant exponential divergence characteristics in a state space for different motion time series or dynamical processes. Note that our method does not need to use either any navigation equipments of ego-vehicle or any high-definition maps. Finally, we conduct extensive experiments on the lane intrusion action recognition dataset called THU-IntrudBehavior, which is collected from real urban roads. The experimental results show that the PSRNet reaches the best accuracy of 98.0\%, which remarkably surpasses classical action recognition approaches by more than 30\%. 
In short, the main contributions in this paper are summarized as follows:
\begin{enumerate}
    \item Based on monocular tracking-by-detection algorithms, we present an object-level lane width normalization strategy such that lane intrusion actions of objects in consecutive video frames can be represented by relative motion time series accordingly. 
    \item A novel lightweight PSRNet that comprises both latent phase space reconstruction layers and a learnable Lyapunov exponent-like classifier is proposed to accomplish classifications of such relative motion time series. 
    \item Experimental results yielded on THU-IntrudBehavior demonstrate that our PSRNet achieves the best performance of 98.0\%.
\end{enumerate}

\section{Related Work}

\subsection{CNN-based Video Action Recognition Approaches} 
In 2014, CNN was at the first time employed for large-scale video classification~\cite{karpathy2014large}. 
In order to learn both spatial and temporal features, D. Tran, \emph{et al}. present C3D~\cite{tran2015learning} through using 3D convolution kernels. 
In the LSTM+CNN approach~\cite{yue2015beyond}, CNN is first utilized to capture spatial features and LSTM is then exploited for temporal feature extraction. 
In 2016, Feichtenhofer \emph{et al}. propose a two-stream fusion network~\cite{feichtenhofer2016convolutional}, where several fusion algorithms are employed to learn corresponding features between pixels. 
In ~\cite{wang2016temporal}, TSN divides video into $k$ parts, then selects those parts randomly, and fuse them by two-stream networks, which solves the problem of being unable to capture long-term features. 
I3D is presented in 2017~\cite{carreira2017quo} to expand 2D convolution kernels into 3D ones and then takes advantage of pre-trained models of image data, which significantly improves performance. But the network architecture must be completely consistent with 2D CNN. 
In 2018, R(2+1)D~\cite{tran2018closer} factorizes 3D convolution filters into separate spatial and temporal components yields and consequently it indeed reduces the amount of parameters. 
Furthermore, Zhou \emph{et al}. propose TRN~\cite{zhou2018temporal} that randomly chooses 2, 3, and 4 frames and then employs MLPs to extract spatial features of images. After making fusion of them, it is capable of having extraction of temporal features more effectively. Note that TRN is usually only suitable for actions with a strong direction. 
SlowFast proposed by Feichtenhofer \emph{et al}. in 2019~\cite{feichtenhofer2019slowfast} contains a high-resolution slow network and a low-resolution fast network. Those two branches can capture local features and global features, respectively. 
Meanwhile, TSM~\cite{lin2019tsm} for video understanding is presented. 
In 2020, Yang \emph{et al}.~\cite{yang2020temporal} propose a temporal pyramid network that the source of features and the fusion of features jointly form a feature hierarchy for the backbone so that it can capture action instances at various tempos.

\begin{figure*}[t]
    \centering
    \includegraphics[width=0.88\textwidth]{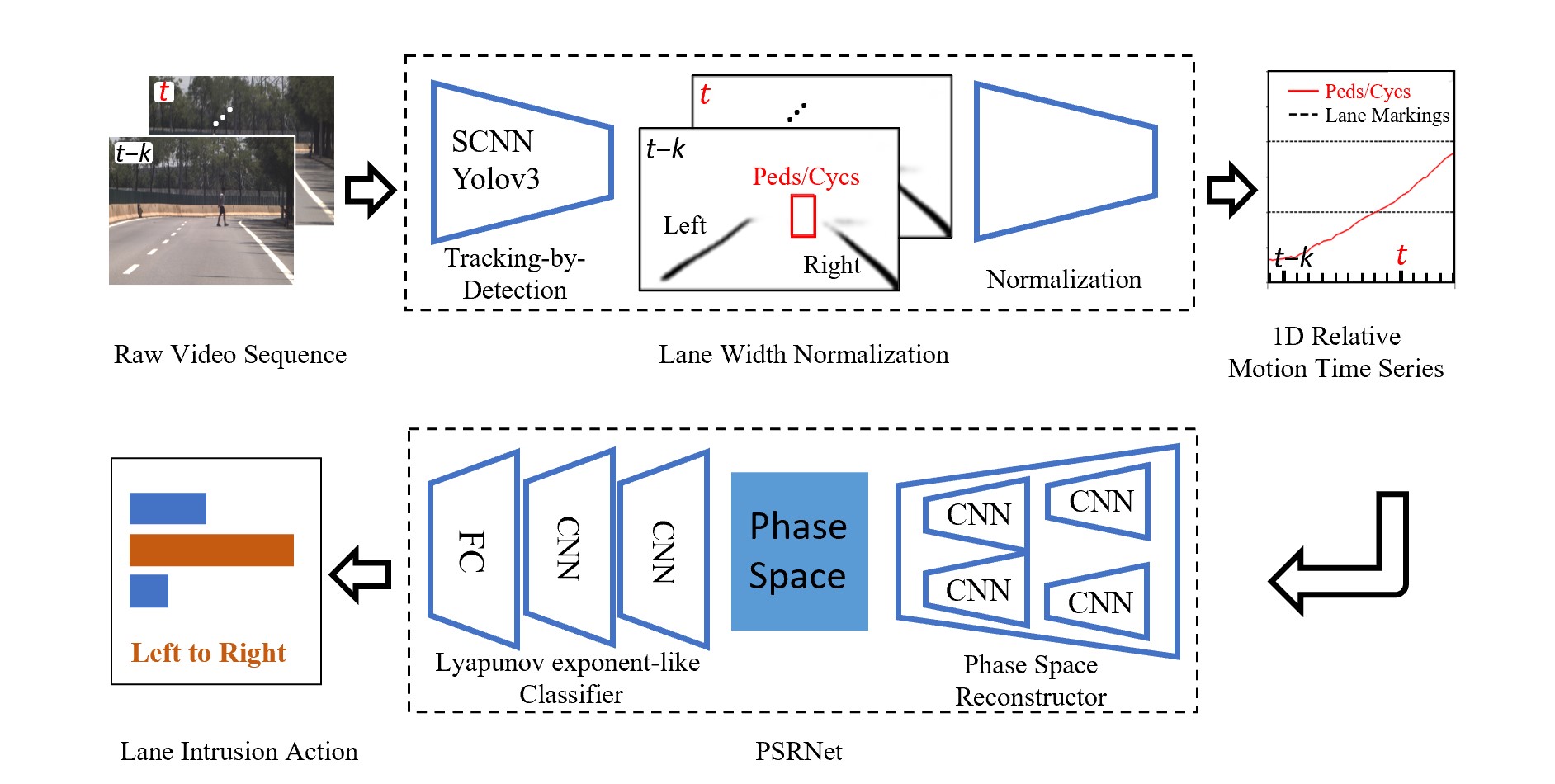}
    \caption{The overall framework of our method.}
    \label{fig:1}
    \vspace{0.1cm}
\end{figure*}

\subsection{Time Series Prediction Based on Deep Learning} 
Recurrent neural networks (RNNs) can handle temporal sequence data. As one of the most representative models in RNN, LSTM~\cite{hochreiter1997long} has competitive performance in long-term dependencies. GRU~\cite{cho2014learning} is a variant of LSTM, which achieves the same performance as LSTM, but it is easier to conduct calculation. 
TCN~\cite{bai2018empirical} is just a temporal CNN model that uses both 1D causal convolution and dilated convolution and is combined with residual networks, which can treat sequence data in parallel and adjust receptive fields flexibly. 
Social GAN~\cite{gupta2018social} introduces a loss that encourages the generative network of GAN to spread its distribution and cover the space of possible paths and designs a new pooling mechanism that is employed to encode subtle cues for all people involved in a scene. 
In~\cite{zhu2019starnet}, Zhu \emph{et al}. propose StarNet that has a star topology with both host network and hub one, where the hub network takes the observed trajectories of all pedestrians and produces a comprehensive spatio-temporal representation of all interactions in crowds. 
In order to integrate human social behaviors in dynamic scenes for prediction tasks, Sun \emph{et al}. propose a recursive social behavior graph~\cite{sun2020recursive}, which is demonstrated to obtain greater expressive power and higher performance. 

\subsection{Tracking-by-Detection Algorithms for Visual Objects and Lane Markings}
In general, visual object detection methods are classified into two-stage and one-stage ones. Two-stage methods including fast R-CNN~\cite{girshick2015fast} and faster R-CNN~\cite{ren2015faster} are able to detect small and even tiny objects effectively but they usually have higher computational complexity, compared to one-stage ones. 
Recently, a couple of sophisticated one-stage models such as Yolo series~\cite{redmon2016you,redmon2017yolo9000,redmon2018yolov3} and SSD~\cite{liu2016ssd} have not only excellent detection performance but also good real-time performance~\cite{liu2019recent}. 
In particular, there are also large-scale datasets for visual object detection tasks like MSCOCO~\cite{lin2014microsoft} and PASCAL VOC~\cite{everingham2010pascal}, which often results in  effective pre-trained models being available.
In fact, visual object detection is the foundation of visual object tracking (VOT), both of which are closely related. 
Based on~\cite{bertinetto2016fully,Danelljan2016Beyond}, Zhu \emph{et al}. present a distractor-aware siamese network for VOT called DaSiamRPN~\cite{zhu2018distractor} that adds RPN and distractor-aware module, and achieves SOTA performance with even a speed of over 160 fps, which model parameters can be updated at runtime.  
Compared to single object tracking (SOT), the research progress of multi-object tracking (MOT) is relatively slow. Actually, DeepSort~\cite{wojke2017simple} proposes a mainstream tracking-by-detection framework, which generally has better performance in the MOT method.
On the other hand, lane markings detection plays an important role in the field of autonomous driving. A collection of large-scale datasets on lane markings detection such as CULane~\cite{pan2018spatial} and Tusimple~\cite{TuSimple} are publicly released. In recent years, as deep learning tremendously proceeds, many lane detection algorithms with exceptional performance like ~\cite{pan2018spatial} and ~\cite{neven2018towards} continue to emerge.

\section{Mothodlogy}

A novel phase space reconstruction network called PSRNet is proposed to accurately recognize lane intrusion actions in high-speed and long-range traffic scenes. 
Fig.~\ref{fig:1} shows the overall framework of our PSRNet. In the upper part of Fig.~\ref{fig:1}, the lane width normalization is employed to transform raw video sequences into relative motion time series. Specifically, raw video sequences are first fed into monocular tracking-by-detection modules to estimate positions of both moving objects, i.e., pedestrians and cyclists, and lane markings and accordingly, resulting positions of objects are then normalized based on pixel lane width so as to generate the motion time series of each object relative to central trajectory of current lane. 
In this case, as shown in the lower part of Fig.~\ref{fig:1}, the PSRNet is proposed to classify the above relative motion time series and eventually accomplish lane intrusion action recognition. Basically, our PSRNet consists of a reconstructor and a classifier, where a reconstructor builds latent phase spaces embedded by motion time series through supervise learning and a classifier learns Lyapunov exponent-like characteristics of phase spaces to make prediction of lane intrusion actions.

\subsection{Lane Width Normalization}
In this subsection, we analyze how to leverage the motion time series to represent the lane intrusion action and discuss the effectiveness of normalization. Especially, we explain why only a monocular camera can be utilized to recognize lane intrusion actions that appear in the real world.

\subsubsection{Position Estimation of Objects}
\begin{figure}[t]
    \centering
    \includegraphics[width=0.78\columnwidth]{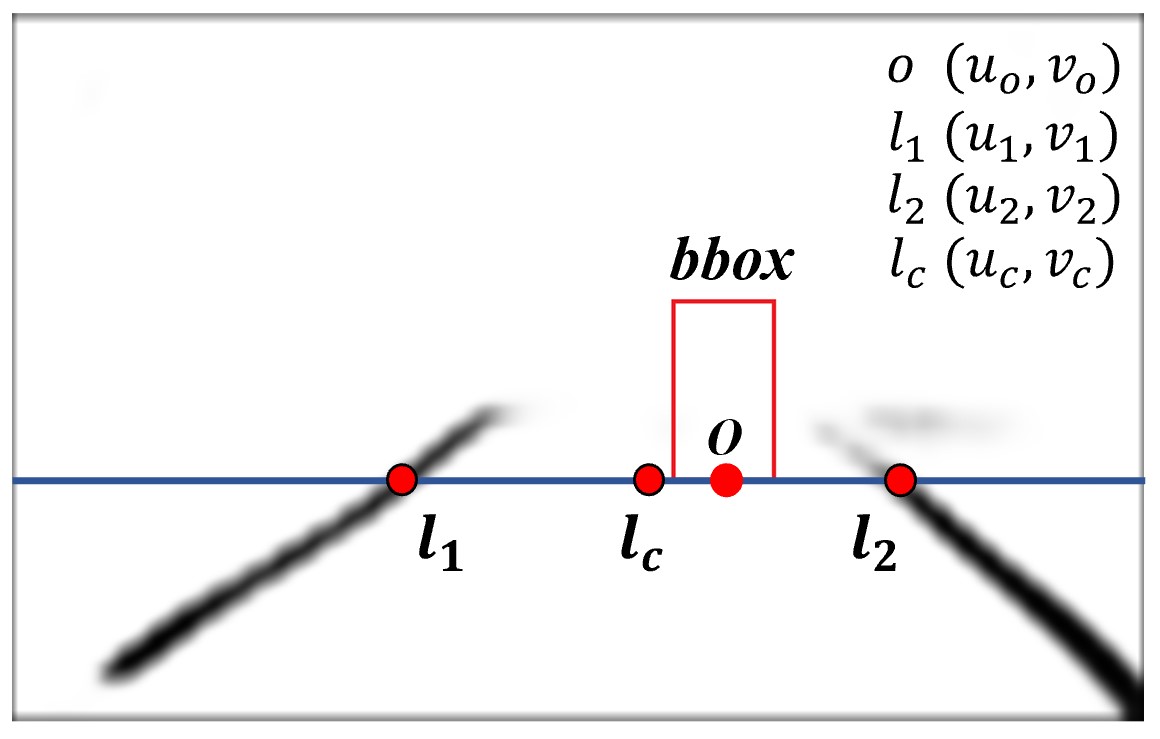}
    \caption{The visual detection results in a unified 1D coordinate system. The black pixel points indicate the lane markings, the red box stands for the bounding box of moving objects, and the blue line expresses the horizontal baseline or the 1D coordinate axis. $l_1$, $l_2$, and $o$ denote positions or 1D coordinates of the left, right lane markings, and the moving object, respectively.} 
    \label{2a}
    \vspace{-0.4cm}
\end{figure}

Using sophisticated deep learning algorithms including visual object and lane markings detections, we can infer bounding boxes of moving objects and pixel segmentations of current left/right lane markings. 
As shown in Fig.~\ref{2a}, the above tracking-by-detection results are all projected onto the identical image coordinate system. A blue line extended along the bottom line of bounding-boxes is referred to as a horizontal baseline of current video frame. 
On the horizontal baseline, we establish a one-dimensional coordinate system in order to clearly mark positions of both moving objects and lane markings. 
As a result, relative motions are evaluated in terms of this coordinate systems. Apparently, it is the same as the $u$-axis in the image coordinate system. Thus the point $o$ denotes the center of objects detected, which 1D coordinates are $(u_{tl} + u_{br}) / 2 $.

\subsubsection{Normalization Based on Lane Width}
Based on deep learning methods for tracking-by-detection, we can locate moving objects $o$ and current left/right lane markings $l_1$ and $l_2$ in the image coordinate system. In order to express the spatial-temporal motion of moving objects, we present an object-level normalization approach. 
Let the intersection $l_c$ between the horizontal baseline and the current lane central trajectory be the origin of the 1D coordinate system. The left side of the origin is defined as negative and the right one as positive. At the same time, the pixel lane width in current video frame is viewed as the normalized unit length. Through normalizing with such a visual lane width in each current frame, the 1D time series of moving objects relative to current lane central trajectory can be calculated. It has 
\begin{align}
    &p_r = \frac{ u_o-u_c }{w} \notag \\
    u_c = (u_1&+u_2)/2, 
    w = \left\vert u_1-u_2 \right\vert
    \label{f_norm}
\end{align}
where $w$ indicates the pixel width of current lanes, $u_c, u_1,$ and $u_2$, respectively, stand for the coordinates of the center point (i.e. the origin), the left and right representatives of current lane markings near moving objects. $o$ denotes the representative of moving objects, and $p_r$ expresses the normalized position of $o$ relative to the central trajectory $l_c$.

\subsubsection{Action Representation by Motion Time Series}
\begin{figure}[t]
    \centering
    \includegraphics[width=0.82\columnwidth]{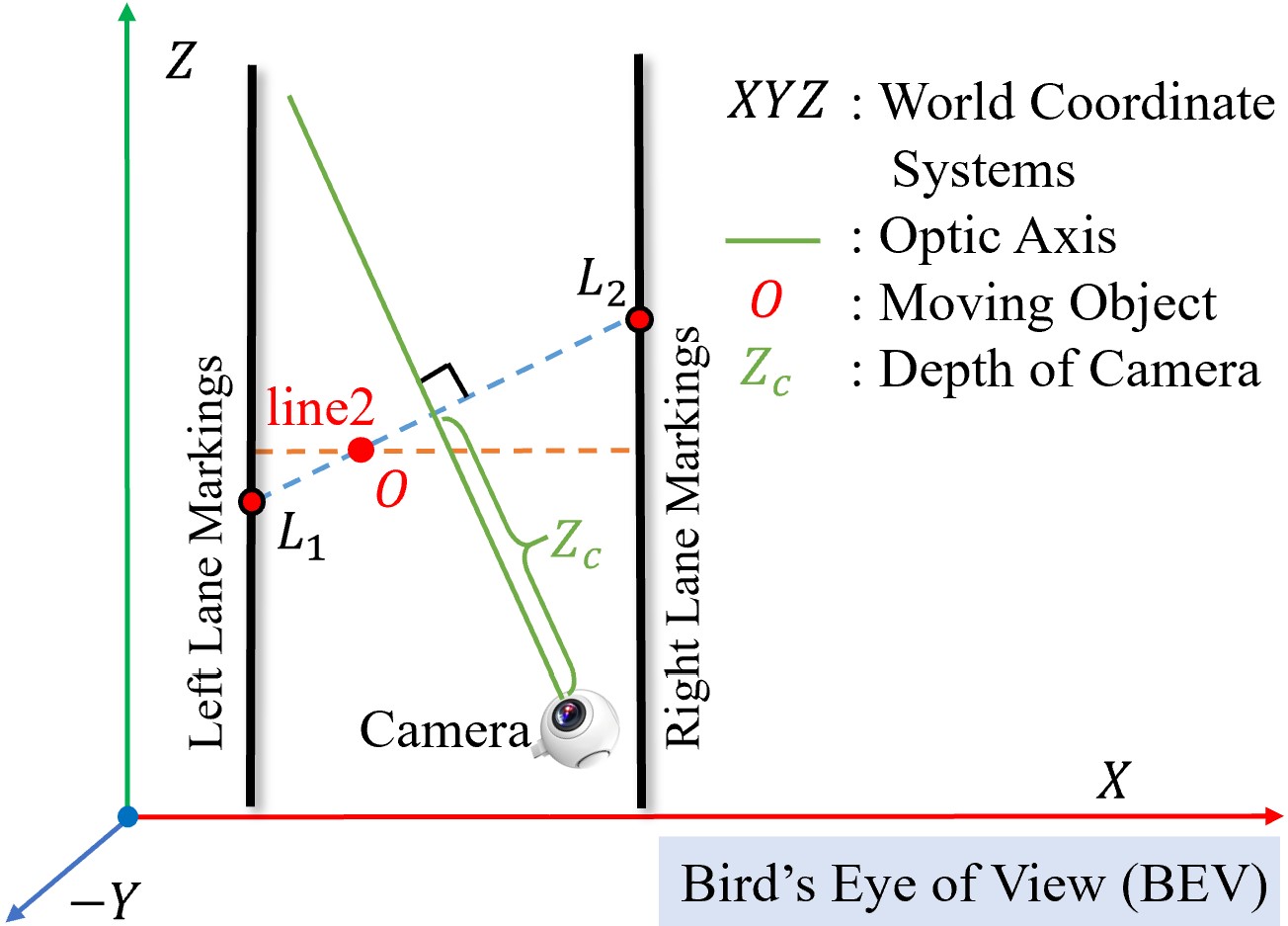}
    \caption{In the world coordinate system, the normalized relative positions are identical for different view angles of camera.} 
    \label{z_c}
    \vspace{-0.3cm}
\end{figure}

In image coordinate, there are the objects $o$ ($u_o, v_o$), lane marking $l_1$ ($u_1, v_o$) and $l_2$ ($u_2, v_o$). Correspondingly, in world coordinate system as shown in Fig.~\ref{z_c}, there are also moving objects $O$ ($X_w^o, Y_w^o, Z_w^o$), left lane markings $L_1$ ($X_w^1, Y_w^o, Z_w^1$), right lane markings $L_2$ ($X_w^2, Y_w^o, Z_w^2$), and camera ($x_c, y_c, z_c$, $\theta_x, \theta_y, \theta_z$). 
The principle of camera imaging geometry, described in Eq. (\ref{project}), illustrates how to project a point ($X_w, Y_w, Z_w$) in the world coordinate system onto a point ($u, v$) in the image ones.

\begin{align}
    \centering
    Z_c \left[\begin{array}{c}
            u \\ v \\ 1
           \end{array}\right] &= 
       \left[\begin{array}{cccc}
        f_x & 0 & u_0 & 0 \\
        0 & f_y & v_0 & 0 \\
        0 & 0 & 1 & 0
       \end{array}\right]
    \left[\begin{array}{cc}
        A & T \\
        0  & 1
       \end{array}\right]
    \left[\begin{array}{c}
        X_w \\ Y_w \\ Z_w \\ 1
       \end{array}\right] \notag \\
    A &= R_z(\theta_z)R_x(\theta_x)R_y(\theta_y), 
    T = [x_c, y_c, z_c]^T 
    \label{project}
\end{align}
where $Z_c$ is the object coordinates in the camera coordinate system, $f_x, f_y, u_0, v_0$ are the camera intrinsic parameters, and the rotation matrix $A$ and the transformation matrix $T$ are the camera external parameters. $R_x$ represents the rotation matrix of rotation angle $\theta_x$, while $(x, y, z)$ denotes the location of camera in the world coordinate system.

Suppose that within a local region, the camera only rotates $\theta_y$ along the $Y$-axis and moves on the $X$-$Z$ plane, and lane markings are parallel to the $Z$-axis. 
As a result, a point in the world coordinate system can be projected onto the one in the corresponding image coordinate system, i.e.,
\begin{small}
\begin{align}
    u = \frac{f_{u1}(X_w)+f_{u2}(Y_w, \theta_y)+f_{u3}(Z_w, \theta_y)+f_{u4}(x_c, z_c)}{Z_c}
    \label{u}
\end{align}
\begin{align}
    v = \frac{f_{v1}(X_w)+f_{v2}(Y_w, \theta_y)+f_{v3}(Z_w, \theta_y)+f_{v4}(x_c, z_c)}{Z_c}
\end{align}
\end{small}where $f$ denotes a linear function. 
Since the monocular camera lacks the depth information, we make a reasonable assumption that the depth ($Z_c$) of both moving objects and their neighboring lane markings is almost the same. So the line where $O$, $L_1$ and $L_2$ are located at is perpendicular to the optical axis, as shown in Fig.~\ref{z_c}. For objects and lane markings, $Z_c$ and $v$ is equal. Thus we can have $Z_w^1$ and $Z_w^2$ below,
\begin{align}
    f_{u1}(X_w^o)+f_{u3}(Z_w^o, \theta) 
   &= f_{u1}(X_w^1)+f_{u3}(Z_w^1, \theta) \notag \\
   &= f_{u1}(X_w^2)+f_{u3}(Z_w^2, \theta)
\end{align}

Then we project the points $O$, $L_1$ and $L_2$ onto the ones in the image coordinate system according to Eq. (\ref{u}). Based on Eq. (\ref{f_norm}), we can get the relative position of objects,
\begin{align}
    p_r = \frac{u_o-(u_1+u_2) / 2}{\left\vert u_1-u_2 \right\vert} 
    \approx  \frac{2 \cdot X_w^o-(X_w^1+X_w^2)}{2 \cdot \left\vert X_w^1-X_w^2 \right\vert}
\end{align}

As shown in Fig.~\ref{z_c}, it is easy to know that the normalized position of points $O$ on the line $L_1 L_2$ is the same as on the line2. It means that relative position $p_r$ is only associated with $X_w^o$ because $X_w^1$ and $X_w^2$ remain fixed. Therefore, lane intrusion actions are independent of object trajectories.

\subsection{PSRNet}
In general, a phase space represents the smallest possible state space of a dynamic system. Each state vector of a dynamic system is expressed as a point in phase space. 
Any time series observations can be regarded as outputs of embedded dynamic systems with unknown system order, which is described by phase state space representation. Accordingly, the evolution process of a dynamic system can be depicted as a motion trajectory from the initial state in phase space. 
Time series is usually analyzed in the time domain. In particular, for the chaotic time series, many studies are carried out in reconstructed nonlinear phase space~\cite{gao2009complex,fan2018short,wang2016temporal}. Therefore, the phase space reconstruction method plays an important role in the processing of chaotic time series. 
In addition, the Lyapunov exponent is defined as the average exponential divergence rate of adjacent motion trajectories in phase space, which is generally viewed as one of several numerical characteristics for identifying nonlinear chaotic motion~\cite{rozenbaum2017lyapunov}.

We devise a PSRNet that comprises both a reconstructor and a classifier. The PSRNet first reconstructs an ensemble of latent phase spaces in the reconstructor and then learns Lyapunov exponent-like characteristics in the classifier.  

\subsubsection{Reconstruction of Latent Phase Space}
\begin{figure}[t]
    \centering
    \includegraphics[width=1\columnwidth]{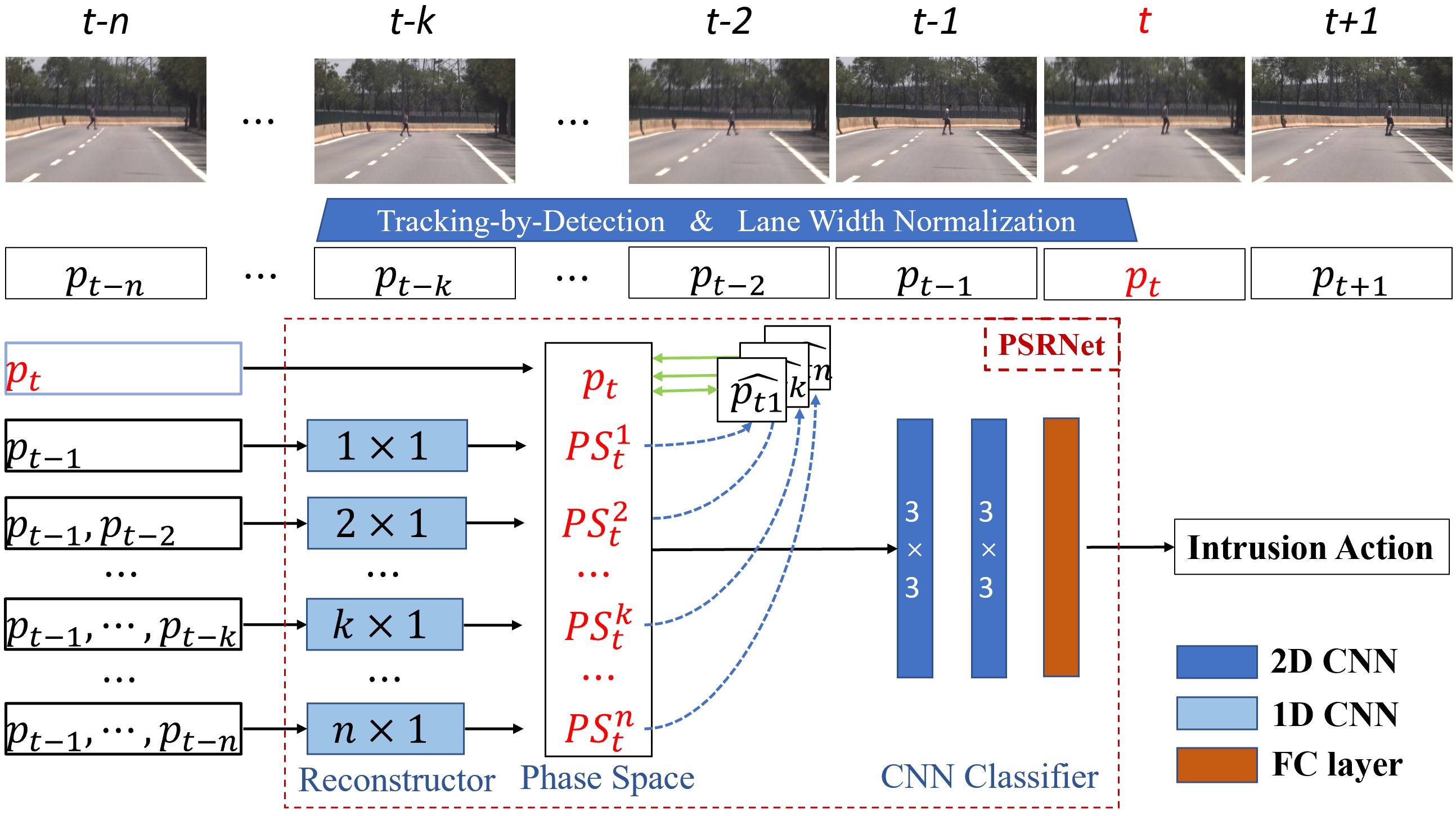}
    \caption{The PSRNet. The reconstructor employs the $n$ 1D-CNNs with different system order to approximate the given time series observations by supervised learning. All the inputs and outputs of such reconstructor are fed into the CNN classifier for lane intrusion action recognition. $n$ is set to $4$ in this work.}
    \label{fig:3}
    \vspace{-0.5cm}
\end{figure}

Based on the idea of traditional phase space reconstruction, the proposed PSRNet builds an $n$-order phase space to represent possible state vectors corresponding to the motion time series. 
As shown in Fig.~\ref{fig:3}, considering that the system order is unknown, the reconstructor in our PSRNet leverages $n$ CNN modules to reconstruct $n$ dynamic systems with different system orders. 
Specifically, the input of each CNN module is the $k$ observations before the current time $t$ in the given time series data $(k = 1, 2, \cdots, n)$, denoted as \{$p_{t-1}$, $\cdots$, $p_{t-k}$\}, so that the $k$-order phase space is described as follows,
\begin{align}
    PS_t^k = \varGamma_k  (p_{t-1}, \cdots, p_{t-k})
\end{align}
where $PS_t^k$ represents the $k$-order phase space at time $t$ and $\varGamma_k$ expresses the end-to-end CNN of order $k$. 
In essence, the reconstructor adopts a supervised learning to accurately approximate nonlinear mappings in nonlinear difference equations with $k$ system orders. 
In other words, the identical time series observations are utilized to learn the approximation of $k$-order difference systems. Note that the observation $p_t$ itself at the current time $t$ in the given time series is employed as the output expectation of CNN modules. Here we adopt the mean square error (MSE) as the loss function of the reconstructor. The loss $\mathcal{L}_{1k}$ of the $k$-order CNN module is
\begin{align}
    \mathcal{L}_{1k} &= \mathcal{F}_{MSE}(\widehat{p_{tk}}, p_t) \\
    \widehat{p_{tk}} &= \mathcal{F}_k (PS_t^k)
\end{align}
where $\mathcal{F}_{MSE}$ indicates the MSE loss function and $\mathcal{F}_k$ expresses the $k$-order linear layer. 

The latent phase space actually expresses an ensemble of $n$ reconstructed dynamic systems. All the inputs and outputs of the CNN modules, i.e. the reconstructor, are aggregated to form the latent phase space of an ensemble of dynamic systems. The ensemble is given below, 
\begin{align}
    PS_t = \mathcal{F}_{cat}(p_t, PS_t^1, \cdots, PS_t^n)
\end{align}
where $\mathcal{F}_{cat}$ stands for the CONCAT operation. 

\subsubsection{Learnable Lyapunov Exponent}
Afterwards, we aggregate all the inputs and outputs of the reconstructor and feed such an ensemble $PS_t$ into another lightweight CNN classifier, whose output is just the classification results of lane intrusion actions. Heuristically, this can be considered as learning to approximate the Lyapunov exponent-like characteristics of reconstructed latent phase spaces $PS_t$ and then making use of them to classify different lane intrusion actions or corresponding motion time series.

Specifically, such Lyapunov exponent-like classifier shown in Fig.~\ref{fig:3} includes two convolutional layers with $3 \times 3$ kernels and one fully-connected layer with softmax, which outputs the final classification scores. The classification results of lane intrusion actions are described as follows:
\begin{align}
    \widehat{l_p} = \mathcal{F}_{c}(PS_t) 
\end{align}
where $\mathcal{F}_{c}$ represents the Lyapunov exponent-like classifier. The loss of such CNN classifier is defined as the cross-entropy loss that is indicated by $\mathcal{L}_2$, i.e., 
\begin{align}
    \mathcal{L}_2 = \mathcal{F}_{CE}(\widehat{l_p}, l_p) 
\end{align}
where $\mathcal{F}_{CE}$ expresses the cross-entropy loss, and $l_p$ denotes the label of lane intrusion actions of \{$p_0$, $\cdots$, $p_t$, $p_{t+1}$, $\cdots$\}. 

In a word, PSRNet employs supervised learning to model the ensemble of hybrid-order nonlinear dynamic systems embedded by time series observations and reconstruct corresponding latent phase spaces. Sequentially, it further learns the Lyapunov exponent-like discrimination of phase spaces to build a classifier in order to enhance the generalization ability of the whole PSRNet. 
The overall loss $\mathcal{L}$ of the PSRNet is given by: 
\begin{align}
    \mathcal{L} = \lambda \cdot \sum_{k=1}^{n} \mathcal{L}_{1k} + \mathcal{L}_2
    \label{loss}
\end{align}
where $\lambda$ indicates the weight coefficient.

\section{Experiments}
\subsection{Data Preparation: THU-IntrudBehavior}
As far as we know, there is no public dataset for lane intrusion action recognition. In order to verify the effectiveness of our PSRNet and initially examine the feasibility of the proposed method in real-world road traffic environments, we built the lane intrusion action recognition dataset called THU-IntrudBehavior. In the data collection scheme, we firmly mounted the U-TRON telephoto lens and the FLIR camera directly in front of the driver's cab to clearly capture road traffic scenes within 150-250m ahead. 
The ego-vehicle speed remains above 70km/h when collecting data. The video data is acquired in three weather conditions of being sunny, cloudy, and rainy, covering various situations. We imitate lane intrusion action on open urban roads and structured roads in the vicinity of the Garden Expo Park, Beijing, China. Intruders or moving objects include pedestrians and cyclists. Such THU-IntrudBehavior is expected to be publicly released very soon. 

After completing data collection, we clean and annotate all the data. We have removal of the incomplete videos and crop relevant video clips in terms of the presence of intrusion behaviors. Furthermore, we accomplish all the annotations including the bounding box labels of pedestrians and cyclists, the pixel-level labels of neighboring lane markings, and the category labels of intrusion behaviors below: 1) intruding into current lane from left to right, 2) intruding into current lane from right to left, and 3) without any lane intrusion. Note that we only focus on lane markings fragments near pedestrians/cyclists in the annotation. 
We got 96 videos online with a total of 10,972 pictures. It contains 154 lane intruders (i.e. pedestrians and cyclists). In order to justly assess the accuracy and robustness of models, we conducted the 3-fold, 5-fold, and 7-fold cross-validation experiments, respectively, in which all the samples must be shuffled randomly.

\subsection{Implementation Details}
\subsubsection{Object Detection and Lane Markings Detection} First, we got ready for the visual object detection model Yolov3~\cite{redmon2018yolov3} pre-trained using the MSCOCO dataset~\cite{lin2014microsoft} and the lane markings detection model SCNN~\cite{pan2018spatial} that was also pre-trained based on the CULane dataset~\cite{pan2018spatial}. Second, we then used the THU-IntrudBehavior to make fine-tuning of the above two pre-trained models, respectively. After those preparations were done, the video frames were continuously fed into both the well-trained Yolov3 and SCNN models. Accordingly, the bounding boxes of moving objects including pedestrians and cyclists and the segmentations of current lane markings for each frame were inferred. Finally, we combined the inferred bounding-boxes and segmentations in a unified coordinate system, and figured out the pixel positions of moving objects and lane markings each frame, in order to give rise to the object-level motion time series of intruders and lane markings.

\subsubsection{Filtering and Normalization} First, we aligned the motion time series data of each pedestrian/ cyclist and removed all false positives using the Hungarian algorithm. Afterwards, we adopted the Kalman filtering to smooth noises in these time series and lane markings data. The initial state values of the Kalman filtering were assumed to the median of the first three values in time series. Finally, the normalized relative motion time series was computed according to Eq. (\ref{f_norm}), all of which would be utilized as inputs of PSRNet.

\subsubsection{PSRNet and Training Details} Specifically, we selected 24 consecutive frames of the motion time series data as inputs to the PSRNet. In the experiment, the maximum system order of latent phase spaces was assumed to be 4. 
In order to reconstruct the phase space, we built 4 one-dimensional CNN modules to form an ensemble, whose input sizes were 1, 2, 3, and 4, respectively. Note that each of 4 CNN modules outputs 8 channels. 
Meanwhile, the Lyapunov exponent-like classifier comprised two convolutional layers with $3 \times 3$ kernel size and one fully-connected layer, where the numbers of channels in the two convolutional layers are 16 and 32, respectively. Furthermore, we select $\lambda = 0.5$ in Eq. (\ref{loss}). In the training algorithm, Adam optimizer was applied to train the model for 100 epochs. The initial learning rate was set to 0.001 with a mini-batch of 32. 

\begin{table}[h]
    \begin{center}
    \caption{Performance comparison with 3-fold\protect\\ cross-validation experiments.}
    \begin{tabular}{l|c} \hline
    Method                            & Accuracy(\%) \\ \hline \hline
    R(2+1)D~\cite{tran2018closer}     & $47.9 \pm 1.5$ \\
    Two-stream~\cite{simonyan2014two} & $52.1 \pm 1.5$ \\
    TRN~\cite{zhou2018temporal}       & $57.3 \pm 3.0$ \\ \hline
    LSTM-based                        & $96.7 \pm 2.5$ \\
    GRU-based                         & $97.4 \pm 1.1$ \\ 
    TCN-based                         & $96.1 \pm 1.7$ \\ 
    MLP-based                         & $95.3 \pm 1.8$ \\ \hline \hline
    \textbf{PSRNet}                   & \textbf{98.0} $\pm$ \textbf{1.6} \\ \hline
    \end{tabular}
    \label{performance}
    \end{center}
    \vspace{-0.3cm}
\end{table}
\subsection{Performance Comparison}
We selected R(2+1)D~\cite{tran2018closer}, the two-stream network~\cite{simonyan2014two}, and TRN~\cite{zhou2018temporal} as the baseline models in this paper. We also compared LSTM-based, GRU-based, TCN-based and MLP-based methods with the proposed method. In this case, all the LSTM-based, GRU-based, and TCN-based models consist of two-layer neural networks with the same number of channels as PSRNet, which inputs are all the above object-level motion time series data. But the MLP-based model has one hidden layer with 512 neurons. In particular, the 3-fold cross-validation experiments for performance evaluation were carried out due to the relatively small number of samples in the THUIntrudBehavior.

The experimental results are listed in Table~\ref{performance}. Compared to 3D CNNs and two-stream networks, the test accuracy of our PSRNet is remarkably improved, while the LSTM-based and GRU-based methods also achieve significant improvements. 
In fact, the PSRNet slightly outperforms the LSTM-based, GRU-based, TCN-based, and MLP-based methods in terms of its lower standard deviation. All the results demonstrate that object-level learning leads to better performance.

\subsection{Ablation Study and Discussion}
In this subsection, we completed extensive ablation experiments in order to analyze the influence of individual components in our PSRNet method. It should be pointed out that all experiments were done on THUIntrudBehavior based on $k$-fold cross-validation ($k=$3, 5, and 7).

\subsubsection{Normalization and Filtering}
\begin{figure}[t]
    \centering
    \begin{tabular}{cc}
        \includegraphics[height=0.43\columnwidth]{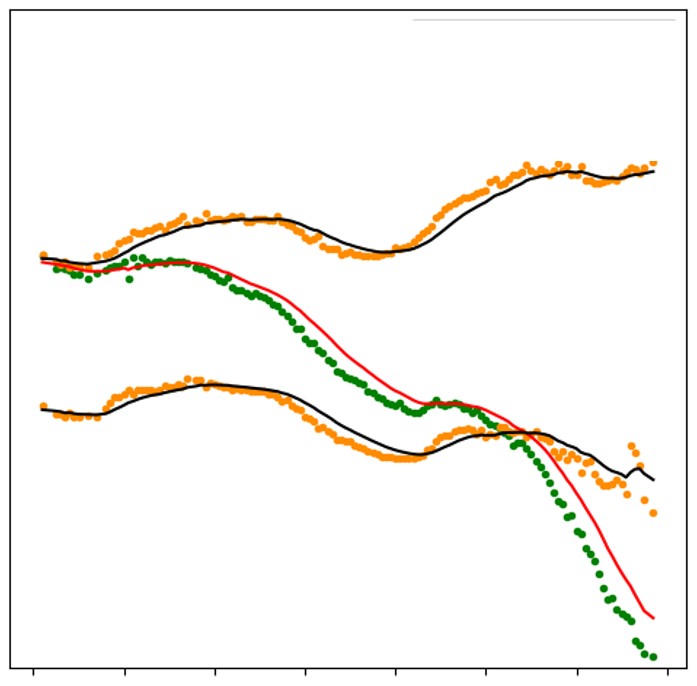}&
        \includegraphics[height=0.43\columnwidth]{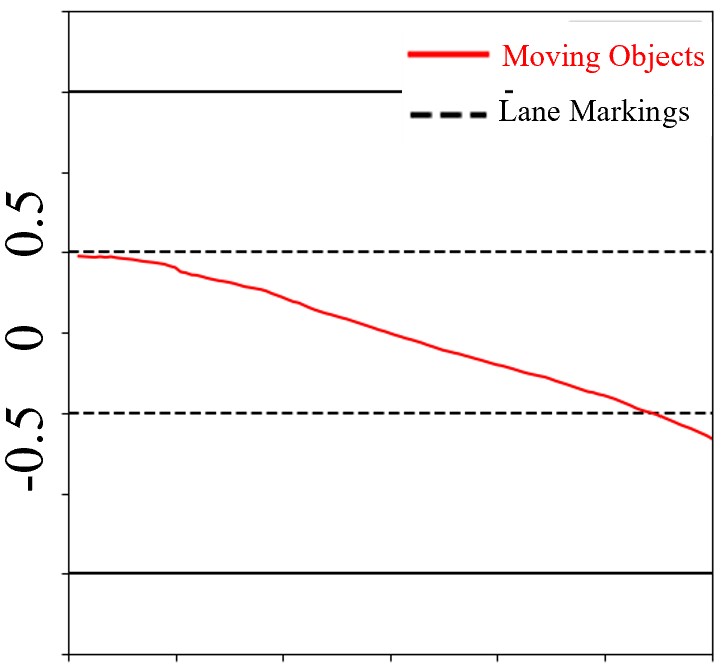}\\
        (a) Before Normalization & (b) After Normalization
        \end{tabular}
    \centering
    \caption{(a) Scattered points indicate position estimations given by visual object detection algorithms. The curves represent trajectories smoothed by the Kalman filtering. (b) The red curve expresses the normalized motion time series of moving objects relative to the central trajectory and the black dashed lines denote current left/right lane markings.}
    \label{5a}
\end{figure}

\begin{table}[h]
    \centering
    \caption{The effects of lane width normalization and filtering.}
    \begin{tabular}{cc|c|c|c} \hline
        Normalization& Filtering   & 3-fold & 5-fold  & 7-fold \\ \hline \hline
                   &            & $65.8 \pm 9.2$ & $65.8 \pm 11.4$ & $50.0 \pm 14.0$ \\
        \checkmark &            & $95.8 \pm 4.1$ & $97.2 \pm 2.8$ & $96.7 \pm 2.5$ \\
        \checkmark & \checkmark & $98.0 \pm 1.6$ & $98.0 \pm 1.7$ & $97.4 \pm 3.3$ \\ \hline
    \end{tabular}
    \label{table2}
\end{table}
We study the effect of filtering and normalization through inputting object-level relative time series into the PSRNet and making comparison of performance, as listed in Table~\ref{table2}. Obviously, without any filtering and normalization, the performance of our PSRNet is basically similar to that of the baselines~\cite{tran2018closer,simonyan2014two,zhou2018temporal}. Both filtering and normalization are crucial to keeping spatial-temporal continuity in object-level relative time series data. 
As shown in Fig.~\ref{5a}(a), the scattered points represent positions given by visual object detection algorithms, while the curves indicate trajectories of pedestrians/cyclists smoothed by the Kalman filtering. The Kalman filtering algorithm is adopted to eliminate outliers and smooth noises, both of which are actually caused by inevitable false positives and missed detections of any tracking-by-detection algorithms. We believe that combination of filtering and normalization simply leads to significant performance improvements. 
Additionally, it is easy to see from Fig.~\ref{5a}(a) that the motion of objects cannot be identified merely through observing the scatter points or the curves. It is even more impossible to determine whether there are intrusion actions. Fig.~\ref{5a}(b) shows the normalized smooth time series relative to the central trajectory of current lane. Because of the normalization of position estimations based on current lane pixel widths each frame, the adverse effect of high-speed movements is reduced greatly. In fact, we can intuitively judge intrusion actions of objects only based on these object-level relative motion time series data.

\begin{table}[h]
    \begin{center}
    \caption{The effects of different system orders of phase spaces.}
    \begin{tabular}{c|c|c|c} \hline
    Order & 3-fold & 5-fold  & 7-fold \\ \hline \hline
      0 & $96.3 \pm 2.3$ & $96.7 \pm 2.7$ & $96.4 \pm 5.7$ \\ \hline
      1 & $97.2 \pm 1.2$ & $96.1 \pm 3.3$ & $96.7 \pm 4.7$ \\
      2 & $97.3 \pm 1.2$ & $96.7 \pm 2.7$ & $96.7 \pm 4.3$ \\
      3 & $97.3 \pm 1.2$ & $97.4 \pm 4.4$ & $97.2 \pm 3.6$ \\ 
      4 & $98.0 \pm 1.6$ & $98.0 \pm 1.7$ & $97.4 \pm 3.3$ \\ \hline
    \end{tabular}
    \label{order}
    \end{center}
\end{table}

\subsubsection{Orders of Latent Phase Spaces} 
In addition, we also performed experiments of the effect of phase spaces with different system orders on the performance of PSRNet. We used zeropadding for lower system orders. As shown in Table~\ref{order}, the accuracy becomes greater as system orders increase. In the process of reconstructing phase spaces, we employed hybrid system orders to capture features of dynamic systems embedded in relative motion time series. Meanwhile, all the inputs and outputs of reconstructed latent phase spaces were aggregated as the input of the classifier so as to further learn Lyapunov exponent-like discriminations. Overall, the application of heuristic knowledge, including phase space and Lyapunov exponent, makes such deep learning models capable of catching inherent dynamic structural features more accurately, which leads to stronger generalization ability.

\subsubsection{Weights in the Loss Function}
The weight coefficient $\lambda$ in Eq. (\ref{loss}) determines the relative contribution of reconstructor and classifier on the whole action recognition task. 
In general, too small weights $\lambda$ will reduce performance due to unbalanced feature representation. 
Therefore, the choice of the weight $\lambda$ needs to empirically be compromised based on trial and error. In all the experiments, we set $\lambda=0.5$.

\section{Conclusions}

This paper proposes a lightweight PSRNet for lane intrusion action recognition based on a monocular camera without using any HD map and precise navigation equipment. First, on the basis of visual tracking-by-detection algorithms for moving objects including pedestrians, cyclists, and lane markings, we present a visual lane width-based normalization policy to generate one-dimensional relative motion time series, which can reliably express lane intrusion actions in video. Second, we build a PSRNet to conduct such motion time series classification. Inspired from classic phase-space reconstruction and Lyapunov exponent methods, the PSRNet comprises several CNN layers and classifier layers, where the former sets up a latent phase space to implicitly represent the movement of pedestrians and cyclists. Thanks to the object-level learning and the introduction of phase space reconstruction and Lyapunov exponent-like classifier, our PSRNet makes it easier to recognize lane intrusion actions in video. The experimental results show that the proposed method even achieves 98.0\% accuracy with high efficiency.

\section*{Acknowledgments}
This work was supported in part by the National Key R\&D Program of China under Grant No.2017YFB1302200 and 2018YFB1600804, by TOYOTA TTAD 2020-08, by a grant from the Institute Guo Qiang (2019GQG0002), Tsinghua University, by Alibaba Group through Alibaba Innovative Research Program, and by ZheJiang Program in Innovation, Entrepreneurship and Leadership Team (2018R01017).

\bibliographystyle{egbibsty}
\bibliography{egbib}

\begin{thebibliography}{10}
\providecommand{\url}[1]{#1}
\csname url@samestyle\endcsname
\providecommand{\newblock}{\relax}
\providecommand{\bibinfo}[2]{#2}
\providecommand{\BIBentrySTDinterwordspacing}{\spaceskip=0pt\relax}
\providecommand{\BIBentryALTinterwordstretchfactor}{4}
\providecommand{\BIBentryALTinterwordspacing}{\spaceskip=\fontdimen2\font plus
\BIBentryALTinterwordstretchfactor\fontdimen3\font minus
  \fontdimen4\font\relax}
\providecommand{\BIBforeignlanguage}[2]{{%
\expandafter\ifx\csname l@#1\endcsname\relax
\typeout{** WARNING: IEEEtran.bst: No hyphenation pattern has been}%
\typeout{** loaded for the language `#1'. Using the pattern for}%
\typeout{** the default language instead.}%
\else
\language=\csname l@#1\endcsname
\fi
#2}}
\providecommand{\BIBdecl}{\relax}
\BIBdecl

\bibitem{gupta2018social}
A.~Gupta, J.~Johnson, L.~Fei-Fei, S.~Savarese, and A.~Alahi, ``{Social GAN}:
  Socially acceptable trajectories with generative adversarial networks,'' in
  \emph{Proceedings of the IEEE Conference on Computer Vision and Pattern
  Recognition}, 2018, pp. 2255--2264.

\bibitem{zhu2019starnet}
Y.~Zhu, D.~Qian, D.~Ren, and H.~Xia, ``{StarNet}: Pedestrian trajectory
  prediction using deep neural network in star topology,'' \emph{arXiv preprint
  arXiv:1906.01797}, 2019.

\bibitem{sun2020recursive}
J.~Sun, Q.~Jiang, and C.~Lu, ``Recursive social behavior graph for trajectory
  prediction,'' in \emph{Proceedings of the IEEE/CVF Conference on Computer
  Vision and Pattern Recognition}, 2020, pp. 660--669.

\bibitem{pellegrini2009you}
S.~Pellegrini, A.~Ess, K.~Schindler, and L.~Van~Gool, ``You'll never walk
  alone: Modeling social behavior for multi-target tracking,'' in \emph{2009
  IEEE 12th International Conference on Computer Vision}.\hskip 1em plus 0.5em
  minus 0.4em\relax IEEE, 2009, pp. 261--268.

\bibitem{lerner2007crowds}
A.~Lerner, Y.~Chrysanthou, and D.~Lischinski, ``Crowds by example,'' in
  \emph{Computer graphics forum}, vol.~26, no.~3.\hskip 1em plus 0.5em minus
  0.4em\relax Wiley Online Library, 2007, pp. 655--664.

\bibitem{karpathy2014large}
A.~Karpathy, G.~Toderici, S.~Shetty, T.~Leung, R.~Sukthankar, and L.~Fei-Fei,
  ``Large-scale video classification with convolutional neural networks,'' in
  \emph{Proceedings of the IEEE conference on Computer Vision and Pattern
  Recognition}, 2014, pp. 1725--1732.

\bibitem{simonyan2014two}
K.~Simonyan and A.~Zisserman, ``Two-stream convolutional networks for action
  recognition in videos,'' in \emph{Advances in neural information processing
  systems}, 2014, pp. 568--576.

\bibitem{lin2019tsm}
J.~Lin, C.~Gan, and S.~Han, ``{TSM: Temporal shift module for efficient video
  understanding},'' in \emph{Proceedings of the IEEE International Conference
  on Computer Vision}, 2019, pp. 7083--7093.

\bibitem{feichtenhofer2019slowfast}
C.~Feichtenhofer, H.~Fan, J.~Malik, and K.~He, ``Slowfast networks for video
  recognition,'' in \emph{Proceedings of the IEEE International Conference on
  Computer Vision}, 2019, pp. 6202--6211.

\bibitem{tran2018closer}
D.~Tran, H.~Wang, L.~Torresani, J.~Ray, Y.~LeCun, and M.~Paluri, ``A closer
  look at spatiotemporal convolutions for action recognition,'' in
  \emph{Proceedings of the IEEE conference on Computer Vision and Pattern
  Recognition}, 2018, pp. 6450--6459.

\bibitem{zhou2018temporal}
B.~Zhou, A.~Andonian, A.~Oliva, and A.~Torralba, ``Temporal relational
  reasoning in videos,'' in \emph{Proceedings of the European Conference on
  Computer Vision (ECCV)}, 2018, pp. 803--818.

\bibitem{yang2020temporal}
C.~Yang, Y.~Xu, J.~Shi, B.~Dai, and B.~Zhou, ``Temporal pyramid network for
  action recognition,'' in \emph{Proceedings of the IEEE/CVF Conference on
  Computer Vision and Pattern Recognition}, 2020, pp. 591--600.

\bibitem{tran2015learning}
D.~Tran, L.~Bourdev, R.~Fergus, L.~Torresani, and M.~Paluri, ``Learning
  spatiotemporal features with {3D} convolutional networks,'' in
  \emph{Proceedings of the IEEE international conference on computer vision},
  2015, pp. 4489--4497.

\bibitem{yue2015beyond}
J.~Yue-Hei~Ng, M.~Hausknecht, S.~Vijayanarasimhan, O.~Vinyals, R.~Monga, and
  G.~Toderici, ``Beyond short snippets: Deep networks for video
  classification,'' in \emph{Proceedings of the IEEE conference on computer
  vision and pattern recognition}, 2015, pp. 4694--4702.

\bibitem{feichtenhofer2016convolutional}
C.~Feichtenhofer, A.~Pinz, and A.~Zisserman, ``Convolutional two-stream network
  fusion for video action recognition,'' in \emph{Proceedings of the IEEE
  conference on computer vision and pattern recognition}, 2016, pp. 1933--1941.

\bibitem{wang2016temporal}
L.~Wang, Y.~Xiong, Z.~Wang, Y.~Qiao, D.~Lin, X.~Tang, and L.~Van~Gool,
  ``Temporal segment networks: Towards good practices for deep action
  recognition,'' in \emph{European conference on computer vision}.\hskip 1em
  plus 0.5em minus 0.4em\relax Springer, 2016, pp. 20--36.

\bibitem{carreira2017quo}
J.~Carreira and A.~Zisserman, ``Quo vadis, action recognition? a new model and
  the kinetics dataset,'' in \emph{proceedings of the IEEE Conference on
  Computer Vision and Pattern Recognition}, 2017, pp. 6299--6308.

\bibitem{hochreiter1997long}
S.~Hochreiter and J.~Schmidhuber, ``Long short-term memory,'' \emph{Neural
  computation}, vol.~9, no.~8, pp. 1735--1780, 1997.

\bibitem{cho2014learning}
K.~Cho, B.~Van~Merri{\"e}nboer, C.~Gulcehre, D.~Bahdanau, F.~Bougares,
  H.~Schwenk, and Y.~Bengio, ``Learning phrase representations using rnn
  encoder-decoder for statistical machine translation,'' \emph{arXiv preprint
  arXiv:1406.1078}, 2014.

\bibitem{bai2018empirical}
S.~Bai, J.~Z. Kolter, and V.~Koltun, ``An empirical evaluation of generic
  convolutional and recurrent networks for sequence modeling,'' \emph{arXiv
  preprint arXiv:1803.01271}, 2018.

\bibitem{girshick2015fast}
R.~Girshick, ``{Fast R-CNN},'' in \emph{Proceedings of the IEEE international
  conference on computer vision}, 2015, pp. 1440--1448.

\bibitem{ren2015faster}
S.~Ren, K.~He, R.~Girshick, and J.~Sun, ``{Faster R-CNN: Towards real-time
  object detection with region proposal networks},'' in \emph{Advances in
  neural information processing systems}, 2015, pp. 91--99.

\bibitem{redmon2016you}
J.~Redmon, S.~Divvala, R.~Girshick, and A.~Farhadi, ``You only look once:
  Unified, real-time object detection,'' in \emph{Proceedings of the IEEE
  conference on computer vision and pattern recognition}, 2016, pp. 779--788.

\bibitem{redmon2017yolo9000}
J.~Redmon and A.~Farhadi, ``Yolo9000: better, faster, stronger,'' in
  \emph{Proceedings of the IEEE conference on computer vision and pattern
  recognition}, 2017, pp. 7263--7271.

\bibitem{redmon2018yolov3}
------, ``Yolov3: An incremental improvement,'' \emph{arXiv preprint
  arXiv:1804.02767}, 2018.

\bibitem{liu2016ssd}
W.~Liu, D.~Anguelov, D.~Erhan, C.~Szegedy, S.~Reed, C.-Y. Fu, and A.~C. Berg,
  ``{SSD: Single shot multibox detector},'' in \emph{European conference on
  computer vision}.\hskip 1em plus 0.5em minus 0.4em\relax Springer, 2016, pp.
  21--37.

\bibitem{liu2019recent}
X.~Liu, Z.~Deng, and Y.~Yang, ``Recent progress in semantic image
  segmentation,'' \emph{Artificial Intelligence Review}, vol.~52, no.~2, pp.
  1089--1106, 2019.

\bibitem{lin2014microsoft}
T.-Y. Lin, M.~Maire, S.~Belongie, J.~Hays, P.~Perona, D.~Ramanan,
  P.~Doll{\'a}r, and C.~L. Zitnick, ``{Microsoft COCO}: Common objects in
  context,'' in \emph{European conference on computer vision}.\hskip 1em plus
  0.5em minus 0.4em\relax Springer, 2014, pp. 740--755.

\bibitem{everingham2010pascal}
M.~Everingham, L.~Van~Gool, C.~K. Williams, J.~Winn, and A.~Zisserman, ``The
  pascal visual object classes (voc) challenge,'' \emph{International journal
  of computer vision}, vol.~88, no.~2, pp. 303--338, 2010.

\bibitem{bertinetto2016fully}
L.~Bertinetto, J.~Valmadre, J.~F. Henriques, A.~Vedaldi, and P.~H. Torr,
  ``Fully-convolutional siamese networks for object tracking,'' in
  \emph{European conference on computer vision}.\hskip 1em plus 0.5em minus
  0.4em\relax Springer, 2016, pp. 850--865.

\bibitem{Danelljan2016Beyond}
M.~Danelljan, A.~Robinson, F.~S. Khan, and M.~Felsberg, ``{Beyond Correlation
  Filters: Learning Continuous Convolution Operators for Visual Tracking},'' in
  \emph{European Conference on Computer Vision}, 2016.

\bibitem{zhu2018distractor}
Z.~Zhu, Q.~Wang, B.~Li, W.~Wu, J.~Yan, and W.~Hu, ``Distractor-aware siamese
  networks for visual object tracking,'' in \emph{Proceedings of the European
  Conference on Computer Vision (ECCV)}, 2018, pp. 101--117.

\bibitem{wojke2017simple}
N.~Wojke, A.~Bewley, and D.~Paulus, ``Simple online and realtime tracking with
  a deep association metric,'' in \emph{2017 IEEE international conference on
  image processing (ICIP)}.\hskip 1em plus 0.5em minus 0.4em\relax IEEE, 2017,
  pp. 3645--3649.

\bibitem{pan2018spatial}
X.~Pan, J.~Shi, P.~Luo, X.~Wang, and X.~Tang, ``Spatial as deep: Spatial cnn
  for traffic scene understanding,'' in \emph{Thirty-Second AAAI Conference on
  Artificial Intelligence}, 2018.

\bibitem{TuSimple}
TuSimple, ``Tusimple benchmark,'' 2017.

\bibitem{neven2018towards}
D.~Neven, B.~De~Brabandere, S.~Georgoulis, M.~Proesmans, and L.~Van~Gool,
  ``Towards end-to-end lane detection: an instance segmentation approach,'' in
  \emph{2018 IEEE Intelligent Vehicles Symposium (IV)}.\hskip 1em plus 0.5em
  minus 0.4em\relax IEEE, 2018, pp. 286--291.

\bibitem{gao2009complex}
Z.~Gao and N.~Jin, ``Complex network from time series based on phase space
  reconstruction,'' \emph{Chaos: An Interdisciplinary Journal of Nonlinear
  Science}, vol.~19, no.~3, p. 033137, 2009.

\bibitem{fan2018short}
G.-F. Fan, L.-L. Peng, and W.-C. Hong, ``Short term load forecasting based on
  phase space reconstruction algorithm and bi-square kernel regression model,''
  \emph{Applied Energy}, vol. 224, pp. 13--33, 2018.

\bibitem{rozenbaum2017lyapunov}
E.~B. Rozenbaum, S.~Ganeshan, and V.~Galitski, ``Lyapunov exponent and
  out-of-time-ordered correlator’s growth rate in a chaotic system,''
  \emph{Physical review letters}, vol. 118, no.~8, p. 086801, 2017.

\end{thebibliography}
\end{document}